\documentclass{article}
\usepackage[preprint,nonatbib]{neurips_2022}
\usepackage[utf8]{inputenc}
\usepackage[T1]{fontenc}

\usepackage{cite}
\usepackage{tikz}
\usepackage{amsmath}
\usepackage{amssymb}
\usepackage{scalerel}
\usepackage{kotex}
\usepackage{multirow}
\usepackage{booktabs}
\usepackage[export]{adjustbox}
\usepackage{tabularx}
\newcolumntype{Y}{>{\centering\arraybackslash}X}
\newcolumntype{N}{>{\centering\arraybackslash}m{.5in}}
\newcolumntype{M}{>{\centering\arraybackslash}m{.47in}}
\newcolumntype{G}{>{\centering\arraybackslash}m{1.7in}}
\usepackage{colortbl}
\usepackage{makecell}
\usepackage[title]{appendix}
\usepackage{placeins}

\usepackage[colorlinks,linkcolor={red},pagebackref=true,breaklinks=true,bookmarks=false]{hyperref}
\usepackage{breakcites}

\usepackage[capitalize]{cleveref}
\crefname{section}{Sec.}{Secs.}
\Crefname{section}{Section}{Sections}
\Crefname{table}{Table}{Tables}
\creflabelformat{equation}{#2\textup{#1}#3}

\title{Real-Time Video Deblurring via Lightweight Motion Compensation}

\newcommand{\name}[1]{\textbf{#1}}

\author{%
\name{Hyeongseok Son}\thanks{Equal contribution.} \hspace{2pt}\thanks{He is currently working at SAIT.} \quad
\name{Junyong Lee}\footnotemark[1] \quad
\name{Sunghyun Cho}\quad
\name{Seungyong Lee}\thanks{Corresponding author.} \\
\\[1pt]
POSTECH\\
\\[1pt]
\{sonhs, junyonglee, s.cho, leesy\}@postech.ac.kr\\
}

\begin{document}

\renewcommand{\thefootnote}{\fnsymbol{footnote}}
\maketitle

\newcommand{\Eq}[1]  {Eq.\ (\ref{eq:#1})}
\newcommand{\Eqs}[1] {Eqs.\ (\ref{eq:#1})}
\newcommand{\Fig}[1] {Fig.\ \ref{fig:#1}}
\newcommand{\Figs}[1]{Figs.\ \ref{fig:#1}}
\newcommand{\Tbl}[1]  {Table \ref{tbl:#1}}
\newcommand{\Tbls}[1] {Tables \ \ref{tbl:#1}}
\newcommand{\Sec}[1] {Sec.\ \ref{sec:#1}}
\newcommand{\SSec}[1] {Sec.\ \ref{ssec:#1}}
\newcommand{\Secs}[1] {Secs.\ \ref{sec:#1}}
\newcommand{\Etal}   {{\textit{et al.}}}
\newcommand{\ie}   {{\textit{i.e.}}}
\newcommand{\eg}   {{\textit{e.g.}}}

\newcommand{\myskip}[1] {}

\newcommand{\setone}[1] {\left\{ #1 \right\}} %
\newcommand{\settwo}[2] {\left\{ #1 \mid #2 \right\}} %

\newcommand{\todo}[1]{{\textcolor{black}{TODO: #1}}}
\newcommand{\son}[1]{{\textcolor{magenta}{hyeongseok: #1}}}
\newcommand{\jy}[1]{{\textbf{\textcolor{blue}{[JY] }}\textcolor{blue}{#1}}}
\newcommand{\sean}[1]{{\textcolor{green}{sean: #1}}}
\newcommand{\sunghyun}[1]{{\textcolor[rgb]{0.6,0.0,0.6}{sunghyun: #1}}}
\newcommand{\change}[1]{{\color{black}#1}}
\newcommand{\changeII}[1]{{\textcolor[rgb]{0.6,0.0,0.6}{#1}}}
\newcommand{\tempch}[1]{{\color{blue}#1}}
\newcommand{\bb}[1]{\textbf{\textit{#1}}}

\renewcommand{\topfraction}{0.95}
\setcounter{bottomnumber}{1}
\renewcommand{\bottomfraction}{0.95}
\setcounter{totalnumber}{3}
\renewcommand{\textfraction}{0.05}
\renewcommand{\floatpagefraction}{0.95}
\setcounter{dbltopnumber}{2}
\renewcommand{\dbltopfraction}{0.95}
\renewcommand{\dblfloatpagefraction}{0.95}

\newcommand{\cm}{\checkmark}

\definecolor{lightlightgray}{gray}{0.96}
\definecolor{royalblue}{RGB}{65,105,225}
\definecolor{royalred}{RGB}{155, 28, 49}
\definecolor{darkgreen}{RGB}{0, 200, 0}
\newcommand*\pct{\protect\scalebox{0.9}{\%}}
\begin{abstract}
While motion compensation greatly improves video deblurring quality, separately performing motion compensation and video deblurring demands huge computational overhead. This paper proposes a real-time video deblurring framework consisting of a lightweight multi-task unit that supports both video deblurring and motion compensation in an efficient way. The multi-task unit is specifically designed to handle large portions of the two tasks using a single shared network and consists of a multi-task detail network and simple networks for deblurring and motion compensation. The multi-task unit minimizes the cost of incorporating motion compensation into video deblurring and enables real-time deblurring. Moreover, by stacking multiple multi-task units, our framework provides flexible control between the cost and deblurring quality. We experimentally validate the state-of-the-art deblurring quality of our approach, which runs at a much faster speed compared to previous methods and show practical real-time performance (30.99dB@30fps measured on the DVD dataset).
\end{abstract}  

\section{Introduction}
\label{sec:intro}
Videos captured in dynamic environments often suffer from motion blur caused by camera shakes and moving objects, which degrades not only the perceptual quality of a video but also the performance of visual recognition tasks~\cite{Ma2016,kupyn2018,Pei2019}. 
To resolve this problem, various video deblurring methods have been proposed and focused on improving deblurring quality while reducing computational costs.

\begin{figure}[t]
\centering
\def \wb {0.499} %
\setlength\tabcolsep{1.2pt}
  \begin{tabular}{@{}cc@{}}
    \includegraphics[width=\wb\textwidth]{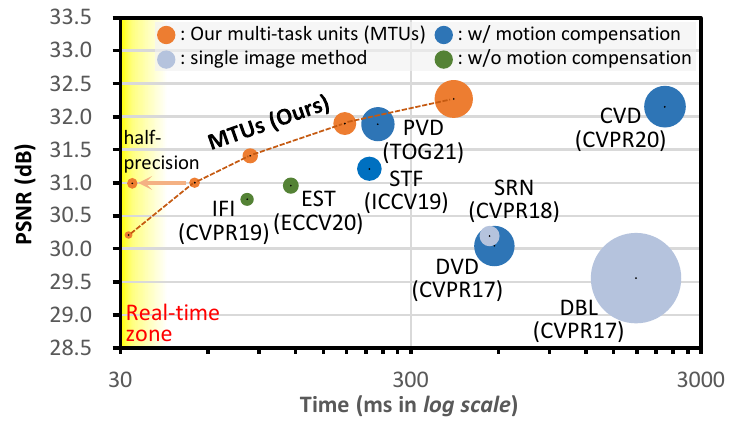} &
    \includegraphics[width=\wb\textwidth]{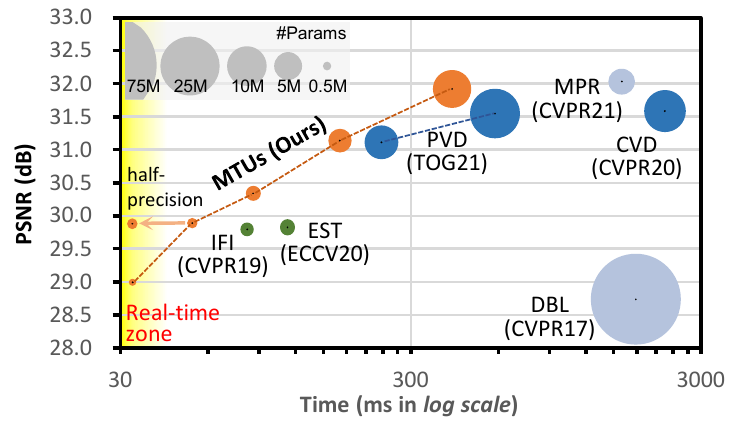}
    \\[-0.02in]
    (a) DVD dataset~\cite{su2017deep} (\cref{tbl:comparison_dvd}) & (b) GoPro dataset~\cite{nah2017deep} (\cref{tbl:comparison_nah}) \\
  \end{tabular}
\caption{Comparison on deblurring efficiency. Our models are indicated in orange circles, each of which is stacked with a different number of multi-task units (MTUs). DBL~\cite{nah2017deep}, SRN~\cite{tao2018scale} and MPR~\cite{Zamir2021MPRNet} are single image motion deblurring methods, and the rests are video deblurring methods. All the running times are measured on the same PC with NVIDIA Geforce RTX 3090 GPU and averaged from 5 runs on each test set containing frames of the 1280$\times$720 resolution. Note that for methods implemented in PyTorch, we wrap the forward function call of a model (\ie, \texttt{torch.nn.Module.forward()}) with CUDA synchronization calls (\ie, \texttt{torch.cuda.synchronize()}) to measure the true inference time taken by the model to deblur a target frame.}
\label{fig:efficiency}
\end{figure}

For successful video deblurring, it is essential to utilize temporal information from neighboring
frames~\cite{cho2012video,delbracio2015,kim2015general}.
For an effective aggregation of temporal information, many approaches align neighboring frames or their features by adopting a separate motion compensation module, such as optical flow~\cite{su2017deep,Zhan2019,Pan2020Cascaded,Son2021PVD,Li2021ARVo}, a spatial transformer network~\cite{kim2018spatio}, dynamic convolution~\cite{Zhou2019ICCV}, and deformable convolution~\cite{Wang2019EDVR}.
While motion compensation improves the deblurring performance as shown in these approaches, heavy computation is required to align blurry frames, resulting in significantly increased computation time, model size, and memory consumption for video deblurring (green vs. blue circles in \cref{fig:efficiency}).
Some approaches pursue efficiency over deblurring quality by omitting the motion compensation module~\cite{kim2017online,Nah2019recurrent,Zhong2020ECCV}.
Nonetheless, these methods can still not perform real-time video deblurring because large network capacity is demanded for aggregating useful temporal information from unaligned blurry frames or features as well as deblurring (green circles in \cref{fig:efficiency}).

This paper proposes an efficient video deblurring framework that reduces the computational overhead but still provides enough performance of motion compensation needed for exploiting temporal information in video frames.
The key component of the proposed framework is a {\em lightweight multi-task unit} 
that handles large portions of both deblurring and motion compensation tasks together using a single shared network.

To maximize the efficiency of our framework, 
both deblurring and motion compensation tasks need to
take full advantage of the shared network in a harmonious way.
However, we found that each task requires different natures of features, which limits the direct sharing of features.
For restoration tasks including deblurring, extracting detail information from a degraded input has shown to be effective in achieving high performance, for which residual (detail) learning has widely been adopted~\cite{Kim2016VDSR,Sajjadi2017,Jo2018,Son2017,su2017deep,Zhou2019ICCV,Wang2019EDVR}.
However, for the motion compensation task,
we observed that structural information of a degraded input is required in addition to the detail information (\cref{ssec:motivation}).

To this end, we specifically design the multi-task unit to improve the compatibility of shared features for video deblurring and motion compensation tasks, and to provide adequate features required for each task.
Our multi-task unit consists of {\em a multi-task detail network}, {\em a deblurring network}, and {\em a motion compensation network},
where the deblurring and motion compensation networks are attached to the shared multi-task detail network.
For the deblurring task, the deblurring network is trained to produce a residual image for deblurring, which induces the multi-task detail network to produce detail features.
On the other hand, the motion compensation task, which requires additional structure information, may hinder the multi-task detail network from producing detail features.
To avoid this, the motion compensation network combines the detail features with pre-computed structure features to constitute structure-injected features, which are then used for the motion compensation task.
As a result, the multi-task detail network is trained to obtain detail features that are shared among the deblurring and motion compensation tasks.

Our multi-task unit has a lightweight structure.
The multi-task detail network has a simple U-Net structure~\cite{Unet15}.
Both deblur and motion networks are composed of only single convolution layers.
For efficient motion compensation,
we use a local correlation-based feature matching operation~\cite{FlowNet,LiteFlowNet}, which does not contain any learnable parameters.
Even though our feature matching-based motion compensation may not be as accurate as motion compensation using a densely computed motion (\eg, an optical flow computed from a large flow estimation model), ours takes a much smaller computational overhead but still significantly improves deblurring performance.

As our multi-task unit is light-weighted, we can stack several units to achieve high deblurring quality without too much computational overhead.
By controlling the number of stacks, we can flexibly control the balance between the model size and deblurring quality.
Thanks to the flexibility of our architecture, our network 
can cover from environments demanding for low-computation to environments where high deblurring quality is desired.

Finally, we demonstrate a real-time video deblurring network with practical performance realized by our multi-task unit.
We thoroughly exploit the trade-off between the deblurring quality and computational cost of our approach.
In the experiment, we found that speed evaluation of existing PyTorch-based image and video deblurring methods do not properly measure the true running time of a model (\SSec{impl_eval}).
To our best knowledge, we are the first to present and perform correct speed evaluations in the field of deblurring.
Each of our model variants shows a much faster running time compared to the corresponding previous methods with similar deblurring quality (orange circles in \cref{fig:efficiency}).

To summarize, our contributions are:
\begin{itemize}
\item A novel lightweight multi-task unit to maximize network sharing between deblurring and motion compensation tasks in a harmonious way;
\item Stacked architecture allowing flexible control on the trade-off between deblurring quality and computational cost;
\item State-of-the-art video deblurring quality with a faster network and practical real-time video deblurring performance.
\end{itemize}

\section{Related Work}
\paragraph{Video Deblurring}
Video deblurring methods focus on taking the benefits of temporal frames.
Classical approaches either directly aggregate temporal information~\cite{matsushita2006,cho2012video,delbracio2015} or solve an inverse problem based on a blur model involving multiple frames~\cite{kim2015general,Wulff2014}.
For an effective aggregation of temporal information, these methods use motion compensation based on a homography~\cite{matsushita2006}, local patch search~\cite{cho2012video}, and optical flow~\cite{delbracio2015,kim2015general}.
However, it is challenging to accurately estimate motion between blurry frames.
For more accurate motion estimation, alternating estimation of motion and deblurred frames has been adopted~\cite{cho2007nonuniform,Bar2007Variational,Wulff2014,kim2015general}, in which a joint energy function for deblurring and motion compensation is iteratively optimized with respect to the blur kernels and motion parameters, respectively.
However, such iterative optimization requires huge computational costs.

\paragraph{Learning-based Video Deblurring using Motion Compensation}
Deep learning-based approaches have been shown to outperform classical methods~\cite{kim2017online,Nah2019recurrent,su2017deep,kim2018spatio,Zhan2019,Zhou2019ICCV,Wang2019EDVR,Wu2020DAVID,Pan2020Cascaded,Zhang2020Recursive,Zhong2020ECCV,Suin2021Gated,Son2021PVD,Li2021ARVo}.
Similar to classical methods, many deep learning-based methods adopt motion compensation for aggregating information from neighboring frames~\cite{su2017deep,kim2018spatio,Zhan2019,Zhou2019ICCV,Wang2019EDVR,Pan2020Cascaded,Son2021PVD,Li2021ARVo}.
Su \Etal~\cite{su2017deep} show that roughly aligning video frames before feeding them into their deblurring network can improve the deblurring quality.
To further improve the deblurring quality, Kim \Etal~\cite{kim2018spatio} and Zhan \Etal~\cite{Zhan2019} introduce optical flow-based motion compensation approaches that align blurry input frames before deblurring.
However, as the optical flow is estimated from blurry frames, its quality is limited, which also limits the deblurring quality.
To address this problem, recently, Son \Etal~\cite{Son2021PVD} propose a pixel volume consisting of multiple matching candidates to robustly handle optical flow errors.
To iteratively improve both motion compensation and deblurring quality, Pan \Etal~\cite{Pan2020Cascaded} introduce a cascaded inference approach.
Their two-stage method is computationally heavy, requiring more than 10$\times$ larger computation time than those of the existing learning-based methods~\cite{Nah2019recurrent,Zhou2019ICCV,Zhong2020ECCV}.
Li \Etal~\cite{Li2021ARVo} extend \cite{Pan2020Cascaded} to resolve optical flow errors by considering all-range volumetric correspondence between the reference and neighboring video frames. However, it also requires a large computational cost similarly to \cite{Pan2020Cascaded}.

Implicit motion compensation methods have also been considered for video deblurring.
Wang \Etal~\cite{Wang2019EDVR} propose deformable convolution networks for motion compensation, while Zhou \Etal~\cite{Zhou2019ICCV} estimate dynamic convolution kernels for both motion compensation and deblurring.
However, although they do not explicitly use motion estimation such as optical flow, these methods require many additional convolution layers with learnable parameters for either estimating deformable or dynamic convolution kernels to handle motion between frames.
Our framework is distinguished from them in that the motion compensation module largely shares the parameters with the deblurring module using multi-task learning.
Thereby, our framework can perform both motion compensation and video deblurring faster with a smaller number of parameters, while showing the deblurring performance comparable to or even better than previous methods.

\paragraph{Learning-based Video Deblurring for Efficiency}
Some approaches pursue efficiency over deblurring quality by omitting the motion compensation module~\cite{kim2017online,Nah2019recurrent,Zhong2020ECCV}.
Kim \Etal~\cite{kim2017online} propose an inter frame-recurrent CNN to sequentially process a blurry video and show that utilizing a deblurred previous frame as an input improves deblurring quality.
Nah \Etal~\cite{Nah2019recurrent} extend \cite{kim2017online} and recurrently process intra-frame features for better propagation of temporal information throughout the inter-frame recurrent pipeline.
It boosts deblurring quality while keeping a lightweight model size but increases the inference time.
Zhong \Etal~\cite{Zhong2020ECCV} use RNN to efficiently extract the spatial features of consecutive frames and aggregate temporal information of the features by channel attention.
While the aforementioned methods are faster than motion compensation-based methods, the deblurring quality is lower because, without motion compensation, it is difficult to aggregate temporal information in video frames that have large motions and blurs.
In addition, all these methods have not provided real-time performance.
Distinguished from these methods, we present a real-time video deblurring framework showing comparable deblurring quality.

\paragraph{Multi-task Learning}
Our method is closely related to multi-task learning in the form of hard parameter sharing.
However, previous approaches~\cite{Thrun1995,Bilen2016,Sener2018multi,Chen2018gradnorm} usually focus on high-level vision tasks, such as classification and scene understanding.
Kokkinos \Etal~\cite{Kokkinos2017} define low-, mid-, and high-level tasks, and show diverse multi-task learning of multi-level tasks.
Still, the low-level tasks in their study include only low-level image recognition tasks, such as boundary, saliency, and normal estimation, but we focus on multi-task learning for the image restoration task.
Song \Etal~\cite{Song2020EdgeStereo} propose multi-task learning of stereo matching and edge detection.
Although they adopt hard parameter sharing, a large number of auxiliary convolution layers is needed for each task.
In contrast, we use only a single convolution layer additionally for each deblurring and motion compensation task.
\input{data/experiments/fig_analysis_feat_compt}
\section{Fast Video Deblurring with Motion Compensation}
\label{sec:method}

\subsection{Improving Compatibility of Shared Features for Multi-Task Learning}
\label{ssec:motivation}
Deblurring and motion compensation tasks are known to benefit each other~\cite{kim2015general,Pan2020Cascaded}.
However, we found that each task requires features that are incompatible in terms of \textit{detail} and \textit{structure}.
Here, following \cite{Pan2018dual,Kim2016VDSR}, we use the terms detail and structure to indicate residual detail computed from a blurry image and structural information of the degraded image, respectively.
For restoration tasks, including deblurring, it is well known from previous approaches~\cite{Kim2016VDSR,Sajjadi2017,Jo2018,Son2017,su2017deep,Zhou2019ICCV,Wang2019EDVR,Lee2021IFAN,Son2021KPAC,Lee2022RefVSR} that residual learning is essential for achieving high performance.
In residual learning, a network focuses on extracting residual information (detail), which is then added to an input degraded image (structure) for producing a deblurred result.

While the deblurring quality improves with the residual learning,
we observed that the motion compensation task requires structural information in addition to the detail information.
For motion compensation, residual details may not provide useful information for matching image patches, such as indistinct contrasts and patterns. 
Consequently, it would be hard to accurately compensate motions between two video frames using only residual details, as shown in Figs.~\ref{fig:SD}b and \ref{fig:SD}f.
By combining structural information with residual details, the accuracy of motion compensation can be improved, as shown in Figs.~\ref{fig:SD}d and \ref{fig:SD}h.

In \cref{ssec:EXP_SD-Separation1}, we validate the different feature characteristics needed for the deblurring and motion compensation tasks.
In the experiment, we compare two networks trained with and without residual learning for both tasks.
The network trained with residual learning produces detail features and shows better deblurring quality, compared to the network trained with non-residual learning that induces features to contain both structure and detail information.
On the other hand, the network trained with non-residual learning shows better motion compensation performance than the network trained with residual learning.

For multi-task learning of the video deblurring and motion compensation tasks that require different natures of features,
we design our multi-task unit to focus on extracting detail features, which are compatible for both deblurring and motion compensation tasks.
Using the detail features, video deblurring can be effectively handled with a simple deblur layer based on residual learning.
However, as shown in \cref{fig:SD}, such detail features cannot be directly used for successful motion compensation.
We propose the structure injection scheme, in which the detail features are combined with structural features that are pre-computed from blurry input frames using a convolution layer, and 
use the combined features for the motion compensation task.
The structure injection scheme for motion compensation improves the compatibility of the shared features among the two tasks, by preventing the motion compensation task from demanding structural information from the multi-task detail network.
In other words, the scheme induces our framework to focus on computing detail features that are useful for the deblurring task and combined with structure features for the motion compensation task to further boost the deblurring quality, as will be quantitatively discussed in \cref{tbl:ablation_embed_table} in \cref{ssec:EXP_SD-Separation2}.

\subsection{Network Overview}
\label{ssec:Network}
\begin{figure}[t]
\centering
\begin{tabular}{cc}
\includegraphics [width=1.0\linewidth] {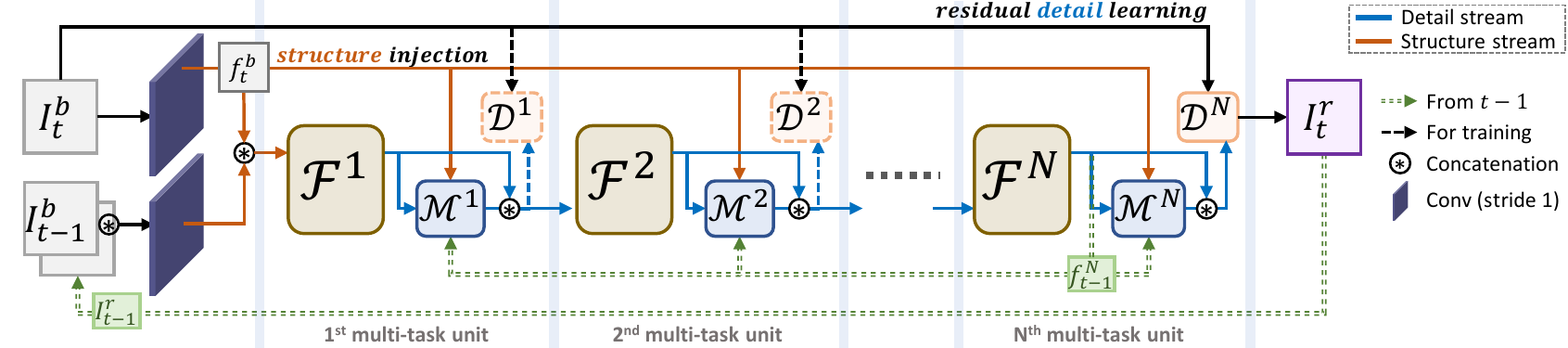}
\end{tabular}
\caption{Overall network architecture. Our network adopts a multi-stacked inter-frame recurrent structure.
The green lines indicate network inputs from the previous iteration $t-1$. The modules connected with dashed single lines are only used during the training.}
\label{fig:multi_stack}
\end{figure}

Our overall network has an iterative structure that stacks $N$ multi-task units for establishing a larger deblurring capability of a network (\cref{fig:multi_stack}).
Before describing the details of the multi-task unit (\cref{ssec:MTU}), we first explain the overall data flow.

Our network adopts an inter-frame recurrent structure.
To deblur a target frame $I_t^b$ at the current time step $t$,
our network receives two consecutive blurry frames $I_{t-1}^{b}$ and $I_t^b$, and the previously restored frame $I_{t-1}^r$ as inputs.
In addition, the network also takes intermediate features computed for the previous frame.
From the input, the network performs deblurring and motion compensation and produces a deblurred resulting frame $I_t^r$.
The network has two convolution layers at its beginning, one of which transforms the current input frame $I_{t}^b$ to a structural feature map ${f}^b_t$,
and the other transforms $[I_{t-1}^b \cdot I_{t-1}^r]$ into a feature space, where $[ \cdot ]$ denotes the concatenation operation along the channel axis.

\begin{figure}[t]
\centering
\begin{tabular}{cc}
\includegraphics [width=1.0\linewidth] {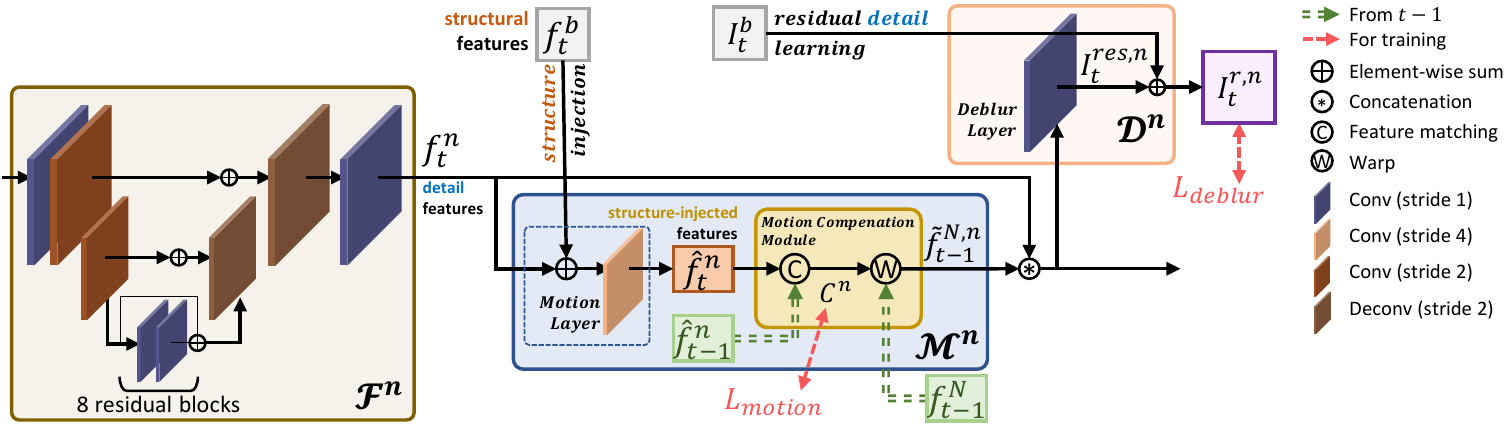}
\end{tabular}
\caption{The $n$-th multi-task unit consists of a multi-task detail network $\mathcal{F}^n$, which is attached with deblurring network $\mathcal{D}^n$ and motion compensation network $\mathcal{M}^n$.}
\label{fig:Network_module}
\end{figure}

\subsection{Multi-task Unit}
\label{ssec:MTU}
Instead of having two separate sub-networks for motion compensation and deblurring tasks,
our multi-task unit computes common detail features shared by two task-specific branches to reduce the computational cost (\cref{fig:Network_module}).
Each multi-task unit consists of three main components:
\textit{multi-task detail network} $\mathcal{F}^n$,
\textit{deblurring network} $\mathcal{D}^n$,
and
\textit{motion compensation network} $\mathcal{M}^n$,
where $n$ is the index of the multi-task unit.
Each multi-task unit shares the same architecture but has different weights.

In the $n$-th multi-task unit, $\mathcal{F}^n$ computes an multi-task detail feature map $f_t^n$ shared by 
$\mathcal{D}^n$
and
$\mathcal{M}^n$.
$\mathcal{F}^n$ is a lightweight network based on the U-Net architecture~\cite{Unet15}, and has most of the learnable parameters of the multi-task unit.
$\mathcal{D}^n$ consists of one convolution layer, named a {\em deblur layer},
that transforms concatenated detail feature maps $f^n_t$ and $\tilde{f}^{N,n}_{t-1}$ of the current and previous frames, respectively, to an output residual image $I_{t}^{res,n}$.
The residual image $I_{t}^{res,n}$ is then added to the input blurry frame $I_t^b$ to produce a restored frame $I_t^{r,n}$.
$\tilde{f}^{N,n}_{t-1}$ is a detail feature map $f^N_{t-1}$ of the previous frame aligned to detail feature map $f^n_{t}$ of the current frame, where we use $\mathcal{M}^n$ for the alignment.
$\mathcal{M}^n$ consists of a single convolution layer, named a {\em motion layer}, and a feature matching-based motion compensation module with no learnable parameters.
In $\mathcal{M}^n$, we first apply the structure injection scheme, where the pre-computed structural feature map ${f}^b_t$ is added to the detail feature map $f_t^n$, which is then transformed to the structure-injected feature map $\hat{f}^n_t$ by the motion layer.
Then, the motion compensation module estimates the motion between the feature maps of the current and previous frames, $\hat{f}^n_t$ and $\hat{f}^n_{t-1}$.
Based on the estimated motion, the module aligns the deblurred feature map $f_{t-1}^N$ of the previous frame to the current target frame and obtains $\tilde{f}^{N,n}_{t-1}$.
We refer the readers to the supplementary material for more detailed network architecture.

Our overall network is light-weighted compared to previous deblurring networks with explicit motion compensation modules~\cite{kim2018spatio,Zhan2019,Pan2020Cascaded,Son2021PVD,Li2021ARVo}, as most of the learnable parameters reside in $\mathcal{F}^n$, and each of $\mathcal{D}^n$ and $\mathcal{M}^n$ has only a single convolution layer.
For notational brevity, we denote the output $I_t^{r,N}$ of the final multi-task unit as $I_t^{r}$ in the rest of the paper.

Our network is trained using two loss functions: a deblurring loss $L_{\scaleto{\textit{deblur}}{5pt}}$, and a motion loss $L_{\scaleto{\textit{motion}}{5pt}}$.
As shown in \cref{fig:Network_module}, we use both loss functions for every multi-task unit when training the network, where all the multi-task units are trained together in an end-to-end fashion.
All the deblurring networks except for $\mathcal{D}^N$ are used only for training and detached after training, as we use only the output of the last deblurring network.
The dashed single lines in \cref{fig:multi_stack,fig:Network_module} illustrate the connections used only in the training phase.

Our model mitigates the different nature of features required by the video deblurring and motion compensation tasks in an effective way.
It adopts long skip-connections (black line denoted as \emph{residual detail learning} in \cref{fig:multi_stack}) between the current target blurry frame $I_t^b$ and output of each deblurring networks $\mathcal{D}^N$.
This promotes the multi-task detail networks to learn detail features, which can improve the deblurring performance.
These detail features may not suffice for improving motion compensation quality, and the additional feature ${f}^b_t$ with structural information is provided to the motion compensation networks.

\subsection{Motion Compensation Network}
\label{ssec:MotionCompensation}

Distinguished from previous works leveraging a heavy motion compensation module~\cite{su2017deep,Zhan2019,Pan2020Cascaded,Son2021PVD,Li2021ARVo}, our motion compensation network is extremely fast, as it shares the network capacity for feature extraction with the deblurring task.
It only requires a single convolution layer (\ie, the motion layer) to constitute structure-injected features and a simple feature matching-based motion compensation module with non-learnable operations (\cref{fig:Network_module}).
We do not include layers to refine the feature matching result for obtaining a motion that is spatially coherent as an optical flow~\cite{FlowNet,LiteFlowNet}, as it is not necessarily required for the motion to be spatially coherent to improve the deblurring quality~\cite{Xue2019video}.

To compute the motion, the module constructs a matching cost volume $C^n_t:\mathbb{R}^{w \times h \times D^2}$ between $\hat{f}^n_t$ and $\hat{f}^n_{t-1}$ based on the cosine-similarity as done in FlowNet~\cite{FlowNet} and LiteFlowNet~\cite{LiteFlowNet}, where $D$ is the maximum displacement.
We then compute a displacement map $W^n_t$ from $C^n_t$ by simply finding the 2D displacement vector corresponding to the largest similarity value for each spatial location in $C^n_t$.
Once $W^n_t$ is computed, we warp the deblurred feature map $f^N_{t-1}$ of the previous frame according to $W^n_t$, and obtain the warped feature map $\tilde{f}^{N,n}_{t-1}$.
The warped feature map $\tilde{f}^{N,n}_{t-1}$ is then concatenated with $f^n_t$ and passed to the next multi-task unit.

For computational efficiency, the motion estimation is performed on downsampled feature maps.
Specifically, the motion layer has stride 4 to compute a downsampled feature map,
which is then used for constructing the cost volume and displacement map.
The displacement map is upsampled using the nearest neighbor interpolation to the original scale, and each displacement vector is scaled as well.
Moreover, for stacked multi-task units (\ie, $n>1$), motion estimation can be omitted for faster inference speed. For example, we may use the displacement map $W^1_t$ computed from the first stack for warping $f^{N}_{t-1}$ in the rest of the stacks. 
Refer to the supplementary material for more details.

\subsection{Loss Functions}
\paragraph{Deblurring Loss}
The deblurring loss $L_{\scaleto{\textit{deblur}}{5pt}}$ is defined using the $L_1$ norm as:
\begin{equation}
    L_{\scaleto{\textit{deblur}}{5pt}} = \frac{1}{M}\sum_{n=1}^{N}\sum_{x=1}^M\lambda_n|I^{r,n}_t(x) - I_{t}^{GT}(x)|,
    \label{eq:loss_aux}
\end{equation}
where $N$ indicates the number of stacks, and $M$ is the number of pixels in a video frame. Here, $x$ denotes a pixel index, $I_{t}^{GT}$ is the ground-truth sharp video frame corresponding to $I_{t}^b$, and $\lambda_{n}$ is a weighting factor for the $n$-th multi-task unit.
In our experiment, we set $\lambda_{n}=0.1$ for $n\in[1,...,N-1]$, and $\lambda_{N}=1$.
Our network could be trained even if the deblurring loss is applied only to the last multi-task unit.
However, applying the deblurring loss to all the intermediate deblurring networks encourages the multi-task detail networks to more effectively learn detail features from the beginning of the stacked network and eventually results in higher-quality deblurring results.

\paragraph{Motion Loss}
The existing video deblurring datasets do not provide ground-truth optical flow labels or displacement maps.
Thus, we instead generate ground-truth cost volumes using ground-truth sharp video frames.
Specifically, for each video frame $I_t^b$, we compute an optical flow map between $I_{t-1}^{GT}$ and $I_{t}^{GT}$ using LiteFlowNet~\cite{LiteFlowNet}.
From the optical flow map, we then generate a ground-truth cost volume $C^{GT}_t$.
The vector at each spatial location of $C^{GT}_t$ is a $D^2$-dim one-hot vector, where the vector element with the value $1$ represents the right match.
Using $C^{GT}_t$, we define the motion loss $L_{\scaleto{\textit{motion}}{5pt}}$ using the cross-entropy as:
\begin{equation}
\begin{split}
    L_{\scaleto{\textit{motion}}{5pt}} =  -\frac{1}{M}\sum_{n=1}^{N}\sum_{x=1}^M\sum_{i=1}^{D^2}{C_t^{GT}(x,i)\log{\left(\textrm{softmax}(C^n_t(x);i)\right)}},
\end{split}
\label{eq:loss_md}
\end{equation}
where $i$ is a channel index, and $C_t^{GT}(x,i)$ and $C_t^{n}(x,i)$ are the feature responses at the spatial location $x$ and channel $i$.
The motion loss encourages the multi-task network to learn multi-task detail features that are useful for motion compensation when it is combined with structural information.
Moreover, the loss trains the motion layer to properly inject structural information into detail features $f^n$ so that $\hat{f}^n$ would be suitable for motion estimation.

The final loss $L_{\scaleto{\textit{total}}{5pt}}$ is defined as: $L_{\scaleto{\textit{total}}{5pt}} = L_{\scaleto{\textit{deblur}}{5pt}}+\alpha L_{\scaleto{\textit{motion}}{5pt}}$, where we use $\alpha=0.01$ in practice.
Thanks to the structure injection scheme in our multi-task unit, $L_{\scaleto{\textit{deblur}}{5pt}}$ and $L_{\scaleto{\textit{motion}}{5pt}}$ assist each other for $\mathcal{F}^n$ to produce multi-task detail features suitable for both deblurring and motion compensation.

\begin{table}[t]
\centering
\aboverulesep = 0.4mm %
\belowrulesep = 0.4mm %
\setlength\tabcolsep{0pt}
\caption{Feature incompatibility of deblurring and motion compensation.}
\scalebox{0.90}{%
    \begin{tabularx}{1.11111\columnwidth}{
    Y Y
    >{\centering}p{0.151\columnwidth}
    >{\centering}p{0.151\columnwidth}
    >{\centering}p{0.151\columnwidth}
    >{\centering\arraybackslash}p{0.151\columnwidth}
    }
    \toprule
    \multirow{4}{*}[-0.15\dimexpr 40\cmidrulewidth]{\makecell{Residual\\learning}} & \multirow{4}{*}[-0.15\dimexpr 40\cmidrulewidth]{\makecell{Motion\\compensation \\module\\+ $L_{motion}$}}& \multicolumn{2}{c}{\multirow{3}{*}{\makecell{Deblurring\\quality}}} & \multicolumn{2}{c}{\multirow{3}{*}{\makecell{Motion \\compensation\\accuracy}}} \\
    && & &  \\
    && & &  \\
    \cmidrule(l){3-6}
    & & PSNR & SSIM & PSNR & SSIM \\
    \midrule
        &     &   29.42 & 0.909 & 26.21 & 0.842 \\
    \rowcolor{lightlightgray}\cm &     &  29.97 & 0.917  & 25.09 & 0.804 \\
    \midrule[0.01mm]
        & \cm & 30.07 & 0.920  & 27.26 & 0.863 \\
    \rowcolor{lightlightgray}\cm & \cm & 29.84 & 0.915 & 27.24 & 0.862 \\
    \bottomrule
    \end{tabularx}%
}
\label{tbl:sdseparation}
\end{table}

\section{Experiments}
\subsection{Implementation Details}
\label{ssec:implementation details}

\paragraph{Dataset}
We use the DVD~\cite{su2017deep} and GoPro~\cite{nah2017deep} datasets to train and validate our network, respectively. 
\change{The DVD dataset provides 71 pairs of blurry and ground truth sharp videos, which contain 6,708 pairs of $1280\times720$ video frames in total.
In our experiments, we use 5,708 frame pairs in 61 videos as our training set and the rest for the test set as done in \cite{su2017deep}.
The GoPro dataset provides 33 video pairs, which contain 3,214 pairs of  $1280\times720$ video frames in total, where the blurry videos are gamma-corrected.}
As done in \cite{nah2017deep}, we use 2,103 frame pairs in 22 videos as the training set and the rest for the test set.

\begin{table}[t]
\centering
\aboverulesep = 0.4mm %
\belowrulesep = 0.4mm %
\setlength\tabcolsep{0mm}

\caption{Ablation study for motion compensation part. Motion compensation module, motion loss $L_{motion}$, and structure injection are components of a motion compensation network $\mathcal{M}$.}

\scalebox{0.90}{%
    \begin{tabularx}{1.11111\textwidth}{
    YYYY
    >{\centering\arraybackslash}p{0.136\textwidth}
    >{\centering\arraybackslash}p{0.136\textwidth}
    >{\centering\arraybackslash}p{0.136\textwidth}
    >{\centering\arraybackslash}p{0.136\textwidth}
    }
    \toprule
    
    \multirow{2}{*}[-0.15\dimexpr 40\cmidrulewidth]{$L_{\scaleto{\textit{deblur}}{5pt}}$} &
    \multirow{2}{*}[-0.15\dimexpr 40\cmidrulewidth]{\makecell{Motion compensation\\ module}} &
    \multirow{2}{*}[-0.15\dimexpr 40\cmidrulewidth]{$L_{\scaleto{\textit{motion}}{5pt}}$} &
    \multirow{2}{*}[-0.15\dimexpr 40\cmidrulewidth]{\makecell{Structure\\injection}} &
    \multicolumn{2}{c}{Deblurring quality} & \multicolumn{2}{c}{Motion compensation accuracy}\\
    \cmidrule(l){5-8}
    & & & & PSNR & SSIM & PSNR & SSIM \\
    
    \midrule
    $\checkmark$ &  & && 29.97 & 0.917 & 25.09 & 0.804  \\
    $\checkmark$ & $\checkmark$ & & & 30.18 & 0.921 & 25.73 & 0.831 \\
    $\checkmark$ & $\checkmark$ & $\checkmark$ & & 29.84 & 0.915 & 27.24 & 0.862  \\
    \midrule[0.01mm]
    $\checkmark$ & $\checkmark$ & $\checkmark$& $\checkmark$ & \textbf{30.30} & \textbf{0.923} & \textbf{27.25} & \textbf{0.864} \\
    \bottomrule
    \end{tabularx}%
}
\label{tbl:ablation_embed_table}
\end{table}

\paragraph{Training Details}
For training, we follow the strategy of \cite{kim2017online}.
From the training set, we randomly select a blurry video to sample random 13 consecutive video frames.
Then, we crop a $256\times256$ patch from each video frame.
We denote the 13 cropped video frames as $\{I_t^b ; t \in [1, ... , 13]\}$.
Then, for each iteration of the training, we feed $I_t^b$, $I_{t-1}^b$, and $I_{t-1}^r$ to the network.
For $t=1$, we set $I_{0}^b = I_1^b$, $I_{0}^r = I_1^b$, and $f_{0}^n = 0$ for $n\in[1, ... , N]$.
In our training, we set the batch size to 8, and 
use the Adam optimizer~\cite{paszke2017automatic} with $\beta_1=0.9$ and $\beta_2=0.999$.
For the analyses of our proposed method using two stacks in \cref{ssec:EXP_SD-Separation},
we trained the models for 300K iterations with a fixed learning rate of $1.0\times10^{-4}$.
For the comparison with previous state-of-the-art methods in \cref{ssec:comparison},
to obtain the best performance, we trained our models for 1,000K iterations with a learning rate of $1.0\times10^{-4}$, 250K iterations with a learning rate of $2.5\times10^{-5}$, and additional 50K iterations with a learning rate of $6.25\times10^{-6}$.
We implemented our models using PyTorch~\cite{paszke2017automatic}.

\subsection{Evaluation}
\label{ssec:impl_eval}
For evaluating deblurring quality, we measure the PSNR and SSIM \cite{wang2004ssim} between deblurring results of the test set and their corresponding ground-truth images.
For evaluating motion compensation quality, as done in \cite{Son2021PVD}, we measure inter-frame alignment accuracy. 
\change{We measure the PSNR and SSIM between $I_{t-1}^{GT}$ and $I_t^{GT}$ after warping $I_{t-1}^{GT}$ using the motion estimated from the last motion compensation module.}

As our motion compensation network estimates motion at the $1/4$ scale, we downsample $I_{t-1}^{GT}$ and $I_t^{GT}$ by 4 for the evaluation.

We also report computation times of deblurring models as they are important measures for practical applicability.
However, we found that the na\"ive measurement in PyTorch does not consider the asynchronous operation of CUDA and yields an inaccurate running time, which is much smaller than the true inference time.
Previous PyTorch-based image and video restoration methods have reported such inaccurate time values, and we remeasure the running time of those methods by wrapping a model with \texttt{torch.cuda.synchronize()} to measure the true running time.
We evaluate our models and previous ones on the same PC with an NVIDIA GeForce RTX 3090 GPU. 

\input{data/experiments/fig_SD_feat_detail_str_injected}

\subsection{Analysis}
\label{ssec:EXP_SD-Separation}

\subsubsection{Feature Incompatibility of Deblurring and Motion Compensation Tasks}
\label{ssec:EXP_SD-Separation1}
We first validate our preliminary observation that video deblurring and motion compensation tasks require different features from the perspective of structure and detail.
For the validation, we compare a stripped-down baseline model and its three variants.
The baseline model is a 2-stacked network (\ie, $N = 2$) trained without residual learning and without motion compensation networks ($\mathcal{M}$).
To the baseline model, we add components one by one to obtain their variants.
The first variant is the baseline model trained with residual learning.
The second one is the baseline model trained with motion compensation module \change{and motion loss $L_{motion}$.
The third one is the second variant model trained with residual learning.}
Note that the structure injection scheme in $\mathcal{M}$ is not included here, and we use the output features of the multi-task detail network directly for the motion compensation task, as we aim to investigate only the feature compatibility between motion compensation and deblurring tasks.
All of the models are trained using the deblurring loss in~\cref{eq:loss_aux}.
\change{For all models, we measure the motion compensation quality using the inter-frame alignment accuracy as in \cref{ssec:impl_eval}. Note that the alignment accuracy of a model without motion compensation module can still be measured using the module, as the module is composed of only non-learnable operations and can be directly applied to the output features of the multi-task detail network.}

\input{data_supp/fig_comparison_MC}

\cref{tbl:sdseparation} shows a comparison of the models, from which we can make the following observations.
\change{First, adopting residual learning improves the deblurring quality but downgrades the inter-frame alignment accuracy (1st vs. 2nd rows of \cref{tbl:sdseparation}).
This indicates that detail-level features learned by residual learning are essential for the deblurring task but are incompatible with the motion compensation task.
Second, employing motion compensation for the model without residual learning improves both the deblurring and motion compensation quality (1st vs. 3rd rows). 
In this case, as there is no residual learning in the model, the entire structure information can be used for improving the motion compensation accuracy. 
Then, the deblurring quality also improves, following the well-known observation that motion compensation helps deblurring.
However, when we na\"ively combine residual learning and motion compensation module (4th row), the multi-task detail network cannot fully exploit neither detail-level nor structural features.
Consequently, the model (4th row) shows downgraded performance for both the deblurring (vs. 2nd row) and the motion compensation (vs. 3rd row) tasks, 
indicating that the two tasks require different characteristics of features. 
To resolve this feature incompatibility, we introduce the structure injection scheme that enables the multi-task detail network to concentrate on exploiting detail-level features.
}

\subsubsection{Ablation Study on Motion Compensation Network}
\label{ssec:EXP_SD-Separation2}
To analyze the effect of our motion compensation approach on deblurring with multi-task learning,
we conduct an ablation study (\cref{tbl:ablation_embed_table}).
All the models in the ablation study are 2-stacked variants of our model that use residual learning for deblurring. 

In \cref{tbl:ablation_embed_table}, merely adopting a motion compensation module of $\mathcal{M}$ increases the deblurring quality even without the motion loss, verifying the usefulness of motion compensation for deblurring.
When the motion loss is simply added to the training of our model without the structure injection scheme of $\mathcal{M}$, it increases the motion compensation quality but degrades the deblurring quality.
In this case, the motion loss induces the multi-task detail network $\mathcal{F}$ to produce features carrying structural information appropriate for motion compensation, but incompatible for deblurring.
However, if we employ the structure injection scheme in $\mathcal{M}$, the deblurring quality significantly increases while the motion compensation quality is retained.
This validates that structure injection plays a key role for the multi-task detail network to focus on producing detail features that can be compatibly shared by both deblurring and motion compensation tasks.

\begin{table}[t]
\centering
\aboverulesep = 0.4mm %
\belowrulesep = 0.4mm %
\setlength\tabcolsep{0mm}
\caption{Quantitative evaluation on the DVD dataset~\cite{su2017deep} (refer to \cref{fig:efficiency}a). The colored dots arrange models showing a similar deblurring quality in PSNR, where each of our models yields an upper bound PSNR at a much faster running time.}
\scalebox{0.70}{%
\begin{tabularx}{1.428\textwidth}{
Y
*{13}{>{\centering}p{0.09051\textwidth}}
>{\centering\arraybackslash}p{0.09051\textwidth}
}
\toprule
& \multirow{2}{*}[-0.2em]{\makecell{DBL\textcolor{orange}{$^{\bullet}$}\\\small{\cite{nah2017deep}}}} %
& \multirow{2}{*}[-0.2em]{\makecell{SRN\textcolor{orange}{$^{\bullet}$}\\\small{\cite{tao2018scale}}}} %
& \multirow{2}{*}[-0.2em]{\makecell{DVD\textcolor{orange}{$^{\bullet}$}\\\small{\cite{su2017deep}}}}%
& \multirow{2}{*}[-0.2em]{\makecell{IFI\textcolor{royalblue}{$^{\bullet}$}\\\small{\cite{Nah2019recurrent}}}} %
& \multirow{2}{*}[-0.2em]{\makecell{EST\textcolor{royalblue}{$^{\bullet}$}\\\small{\cite{Zhong2020ECCV}}}}
& \multirow{2}{*}[-0.2em]{\makecell{STF\textcolor{green}{$^{\bullet}$}\\\small{\cite{Zhou2019ICCV}}}} %
& \multirow{2}{*}[-0.2em]{\makecell{PVD\textcolor{cyan}{$^{\bullet}$}\\\small{\cite{Son2021PVD}}}}
& \multirow{2}{*}[-0.2em]{\makecell{CVD\textcolor{purple}{$^{\bullet}$}\\\small{\cite{Pan2020Cascaded}}}}
& \multicolumn{4}{c}{Ours$^{n}$ ($n$: \texttt{\#} of stack)}
& Ours$^n_{hp}$
& Ours$^n_L$ \\
\cmidrule(lr){10-13}
\cmidrule(l){14-15}
&&&&& &  &   & & $n=1$\textcolor{orange}{$^{\bullet}$} & 2\textcolor{royalblue}{$^{\bullet}$} & 4\textcolor{green}{$^{\bullet}$} & 10\textcolor{cyan}{$^{\bullet}$} & 2\textcolor{royalblue}{$^{\bullet}$} &10\textcolor{purple}{$^{\bullet}$}\\
\midrule

PSNR & 29.56 & 30.20 & 30.05 & 30.75 & 30.97 &31.21 &  31.89 & 32.15 & 30.21 & 31.00 & 31.41 & 31.90 & 30.99 & \textbf{32.27} \\
SSIM & 0.913 & 0.922 & 0.920 & 0.930 & 0.931 & 0.935 & 0.944 & \textbf{0.949} & 0.921 & 0.933 & 0.937 & 0.943 & 0.932 & 0.947 \\
\midrule[0.01mm]
Params (M) & 75.92 & 3.76 & 15.31 & 1.67 & 2.23 & 5.37 & 10.51 & 16.19 & \textbf{0.54} & 1.04 & 2.04 & 5.03 & 1.04 & 13.80 \\
Time (ms) & 1790& 560& 581& 82 & 116 & 216 & 231 & 2250 & \textbf{32} & 54 & 84 & 178 & \textbf{33} & 421 \\
\bottomrule
\end{tabularx}%
}
\label{tbl:comparison_dvd}
\end{table}
\input{data_supp/fig_comparison_DVD_1}
\cref{fig:SD_feat} visualizes samples of detail and structure-injected feature maps in our model variants with and without the structure injection scheme.
In the model without structure injection, the multi-task detail network produces the features where the mixture of structural and detail information exists (the first column of the figure).
In the model, the features are used for both deblurring and motion compensation tasks, eventually decreasing deblurring performance.
On the other hand, in our final model with the structure injection scheme, the multi-task detail network can produce shared features $f_t^n$ containing only detail information suitable for residual learning of deblurring (the second column of the figure). At the same time, structure-injected features $\hat{f}_t^n$ have both structure and detail information suitable for motion compensation (the last column).

\subsubsection{Effectiveness of Lightweight Motion Compensation Network Compared with Previous Approaches}
While motion compensation improves the deblurring performance as shown in many previous approaches~\cite{su2017deep,Zhan2019,kim2018spatio,Zhou2019ICCV,Wang2019EDVR,Pan2020Cascaded,Son2021PVD,Li2021ARVo}, heavy computation is required to align blurry frames, resulting in significantly increased computation time, model size, and memory consumption for video deblurring.
However, 
as our proposed multi-task unit handles large portions of both deblurring and motion compensation tasks using a single shared network, 
the motion compensation task requires only a small computational overhead but still provides enough performance needed for exploiting temporal information in video frames.

\begin{table}[t]
\centering
\setlength\tabcolsep{0mm}
\aboverulesep = 0.4mm %
\belowrulesep = 0.4mm %
\caption{Quantitative evaluation on the GoPro dataset~\cite{nah2017deep} (refer to \cref{fig:efficiency}b).}
\scalebox{0.7}{%
\begin{tabularx}{1.428\textwidth}{
Y
*{11}{>{\centering}p{0.106\textwidth}}
>{\centering\arraybackslash}p{0.106\textwidth}}

\toprule
& \multirow{2}{*}[-0.2em]{\makecell{DBL\textcolor{orange}{$^{\bullet}$}\\\small{\cite{nah2017deep}}}} &
\multirow{2}{*}[-0.2em]{\makecell{IFI\textcolor{royalblue}{$^{\bullet}$}\\\small{\cite{Nah2019recurrent}}}} &
\multirow{2}{*}[-0.2em]{\makecell{EST\textcolor{royalblue}{$^{\bullet}$}\\\small{\cite{Zhong2020ECCV}}}} &
\multirow{2}{*}[-0.2em]{\makecell{PVD\textcolor{cyan}{$^{\bullet}$}\\\small{\cite{Son2021PVD}}}} &
\multirow{2}{*}[-0.2em]{\makecell{PVD$_{L}$\textcolor{purple}{$^{\bullet}$}\\\small{\cite{Son2021PVD}}}} &
\multirow{2}{*}[-0.2em]{\makecell{CVD\textcolor{purple}{$^{\bullet}$}\\\small{\cite{Pan2020Cascaded}}}} & \multicolumn{4}{c}{Ours$^n$ ($n$: \texttt{\#} of stacks)} & Ours$^n_{hp}$ & Ours$^n_L$ \\
\cmidrule(lr){8-11} \cmidrule(l){12-13}
&  &  &  &  &  &  & $n=1$\textcolor{orange}{$^{\bullet}$} & 2\textcolor{royalblue}{$^{\bullet}$} & 4 & 10\textcolor{cyan}{$^{\bullet}$} & 2\textcolor{royalblue}{$^{\bullet}$}& 10\textcolor{purple}{$^{\bullet}$} \\
\midrule
PSNR & 28.74 & 29.80 & 29.83 & 31.11 & 31.55 & 31.59 & 29.00 & 29.91 & 30.34 & 31.14 & 29.90 & \textbf{31.92} \\
SSIM & 0.912 & 0.926 & 0.924 & 0.943 & 0.948 & \textbf{0.952} & 0.915 & 0.928 & 0.933 & 0.942 & 0.928 & 0.950 \\
\midrule[0.01mm]
Params (M) & 75.92 & 1.67 & 2.23 & 10.51 & 23.36 & 16.19 & \textbf{0.54} & 1.04 & 2.04 & 5.03 & 1.04 & 13.80 \\
Time (ms) & 1788 & 83 & 113 & 238 & 585 & 2241 & \textbf{33} & 53 & 86 & 171 & \textbf{33} & 416 \\
\bottomrule
\end{tabularx}%
}
\label{tbl:comparison_nah}
\end{table}
\input{data_supp/fig_comparison_GoPro_1}

To validate the effectiveness of our lightweight motion compensation network $\mathcal{M}$, we compare ours with motion compensation networks utilized in previous video deblurring methods, PVD \cite{Son2021PVD} and STF~\cite{Zhou2019ICCV}.
In PVD, a separate flow estimation network (BIMNet) is utilized to obtain an optical flow between neighboring and target frames.
The flow is then used to construct a pixel volume that comprises multiple candidate pixels of a neighboring frame matched to each pixel of a target frame.
The pixel volume is directly fed to the deblurring network and significantly improves the deblurring quality.
In STF, the filter adaptive convolution network (FAN) is embedded within the deblurring network and estimates dynamic per-pixel convolution kernels. 
The per-pixel kernels are then convolved to the features of a neighboring frame for aligning them to the current target features.

For the validation,
we compare inference time and alignment results of motion compensation networks.
For visualizing alignment results of BIMNet~\cite{Son2021PVD} and our motion compensation network $\mathcal{M}$,
we first predict the motion from $I^b_{t}$ to $I^b_{t-1}$.
Then, using the motion, we directly warp the previous sharp frame $I^{GT}_{t-1}$ for aligning it to the current frame, as described in \cref{fig:SD}.
However, for FAN~\cite{Zhou2019ICCV}, 
it is not straightforward to apply motion compensation to a frame in the image domain because FAN does not predict a flow containing per-pixel displacement between temporal features. 
Instead, FAN implicitly performs motion compensation in the feature space by convolving per-pixel kernels to the features of the previous frame.
Hence, for FAN, we visualize samples of the resulting features aligned to features at the current time step, as done in~\cite{Zhou2019ICCV}.

\cref{fig:MC_comparison} shows the results.
For temporal frames exhibiting small motion (the top row of \cref{fig:MC_comparison}), all compared methods properly compensate the motion between the current and previous frames.
For frames with larger motion (the bottom row), while BIMNet and ours well align the previous frame to the current frame, FAN fails to do so.
This is mainly due to the limited spatial size of the dynamic kernels of FAN utilizing $5\times5$ kernels in practice (corresponds to our motion compensation module with $D=2$), which is not enough for handling large motion between temporal features.
It is also difficult for FAN to adopt a larger kernel size because the cost would become intractable to manage per-pixel kernels and their convolution operations.
Distinguished from BIMNet and FAN that require heavy computations for maintaining a large motion estimation network and for employing per-pixel adaptive convolution operations, respectively, ours requires only a single convolution layer (\ie, motion layer) and a simple local feature matching operation for the motion compensation task.
It is noteworthy that our approach still produces moderately motion-compensated results, with $158\times$ and $68\times$ smaller model size than those of BIMNet and FAN, respectively, and $18\times$ faster inference speed.

\input{data_supp/fig_comparison_real_1}
\subsection{Comparison with Previous Methods}
\label{ssec:comparison}
Finally, we compare our method with state-of-the-art single image~\cite{nah2017deep,tao2018scale} and video deblurring methods~\cite{su2017deep,Nah2019recurrent,Zhou2019ICCV,Zhong2020ECCV,Pan2020Cascaded,Son2021PVD}.
The most recent method~\cite{Li2021ARVo} is not included in this comparison since the source code is not publicly available.

For comparison, we evaluate our methods with the ones on the DVD dataset \cite{su2017deep} and the GoPro dataset~\cite{nah2017deep}.
The compared models are trained with the training set in the corresponding dataset.
We compare our models, each of which is stacked with 1, 2, 4, and 10 multi-task units, denoted by Ours$^n$, where $n$ indicates the number of stacks.
In addition, considering the trade-off between computational costs and deblurring performance, we also include our 10-stacked model with larger parameters (Ours$^{10}_L$), for which we increase the number of channels in the convolution layers from 26 to 48.
We also include the performance of our 2-stacked model (Ours$^{2}_{hp}$) implemented in half-precision~\cite{micikevicius2018mixed}, which also shows the real-time deblurring speed with practical deblurring performance.

As \cref{fig:efficiency} indicates, our multi-stacked models show higher PSNRs than all the other methods at similar inference speeds.
\cref{tbl:comparison_dvd} shows a quantitative comparison on the DVD dataset~\cite{su2017deep}, respectively.
Ours$^2$ outperforms the single image methods~\cite{nah2017deep,tao2018scale}, DVD~\cite{su2017deep} and IFI~\cite{Nah2019recurrent}, while it has a significantly smaller model size and higher speed.
Ours$^{2}_{hp}$ runs in real-time speed while showing similar deblurring quality with IFI and EST, which are designed primarily for efficiency.
Compared to EST~\cite{Zhong2020ECCV} and STF~\cite{Zhou2019ICCV}, Ours$^4$ achieves a higher deblurring quality at a higher speed.
Compared to PVD~\cite{Son2021PVD}, Ours$^{10}$ runs at a faster speed with smaller model size, while achieving comparable deblurring performance.
Compared to CVD~\cite{Pan2020Cascaded}, Ours$^{10}$ achieves slightly lower deblurring performance, but it shows $10\times$ faster speed with $3\times$ smaller model size.
Ours$^{10}_L$ achieves higher PSNR than CVD with a much faster speed, showing the efficiency of our approach.
CVD shows high deblurring accuracy, but its inference is much slower than others.
The slow computational time is due to a large model, including motion compensation and deblurring networks, and iterative utilization of the large model for processing a single frame.
Specifically, for deblurring a single frame, CVD needs to align eight pairs of frames leveraging the flow estimation method~\cite{Sun2018PWC}, which is already computationally heavier (8.75M parameters) than previous deblurring methods.

\cref{tbl:comparison_nah} shows a quantitative comparison on the GoPro dataset~\cite{nah2017deep}.
We compare our models with previous methods, DBL~\cite{nah2017deep}, IFI~\cite{Nah2019recurrent}, EST~\cite{Zhong2020ECCV}, CVD~\cite{Pan2020Cascaded}, and PVD~\cite{Son2021PVD}, for which models pre-trained with the dataset are available.
As the table shows, Ours$^{2}_{hp}$ with real-time speed still achieves comparable PSNR with both IFI and EST, while Ours$^{4}$ outperforms them. 
Ours$^{10}$ also shows comparable PSNR to PVD with faster speed.
It is worth noting that Ours$^{10}_L$ outperforms the large-size model of PVD (PVD$_L$) in PSNR with faster speed, implying a better trade-off between deblurring quality and computational cost.
Ours$^{10}_L$ also shows higher PSNR than CVD despite its smaller computational cost.

\cref{fig:comparison_DVD_1}, \cref{fig:comparison_GoPro_1}, and \cref{fig:comparison_real_1} show qualitative comparisons on the DVD dataset~\cite{su2017deep}, GoPro dataset~\cite{nah2017deep}, and real-world blurred videos~\cite{cho2012video}.
Due to the severe blur in the input frames, the results of the other methods still have remaining blur.
On the other hand, the results of our 10-stacked model show better-restored details.
In addition, Ours$^{2}_{hp}$ removes the majority of blurs successfully, comparable to efficiency-pursuing methods such as IFI and EST, despite its real-time speed.
We found that our-real time model can also improve the performance of object detection on blurry scenes successfully when the model is used as pre-processing. It shows the high applicability of our model to various real-time applications.
Additional results and the object detection examples can be found in the supplementary material.
\section{Conclusion}

We proposed a lightweight multi-task unit that supports both video deblurring and motion compensation tasks.
For cost-effective incorporation of motion compensation into video deblurring,
we design the multi-task unit to maximally share the network capacity by resolving the different natures of features demanded by each task.
The multi-task unit maintains streams of detail features commonly required for both tasks and additionally injects pre-computed structure features for motion compensation.
Our approach reduces the computational overhead, especially for motion compensation, and enables real-time video deblurring.
By simply stacking multiple multi-task units, our deblurring framework can flexibly control the trade-off between deblurring quality and computational costs,
achieving state-of-the-art performance with faster speed.

While structure injection plays a key role in improving motion compensation and deblurring quality, structural information is currently computed with a simple convolution layer.
We plan to focus on extracting better structural features while suppressing computational costs.
The improved structural features may be leveraged to boost motion compensation quality and enhance residual features, promoting better deblurring performance.

\vspace{-8pt}
\subsection*{Acknowledgments}
This work was supported by the Ministry of Science and ICT, Korea, through 
IITP grants
    (SW Star Lab, 2015-0-00174;
    AI Innovation Hub, 2021-0-02068;
    Artificial Intelligence Graduate School Program (POSTECH), 2019-0-01906)
and NRF grant
    (2018R1A5A1060031).

\bibliographystyle{splncs04} 
\bibliography{ms}       

\begin{appendices}
\section{Analysis on Motion Compensation Network}
\subsection{Implementation Detail of Motion Compensation Module}
\label{sec:mc_detail}
The motion compensation network estimates motion between the features $\hat{f}^n_{t}$ and $\hat{f}^n_{t-1}$, which are the modulated structure features resulting from a motion layer.
Inspired by the correlation layer of FlowNet~\cite{FlowNet}, the network first constructs a matching cost volume based on the cosine similarity between the features.
Specifically, given structure-injected features $\hat{f}_{t}^{n}$ and  $\hat{f}_{t-1}^{n}$, the cosine similarity between the two features at $x_1$ in $\hat{f}_{t}^{n}$ and $x_2$ in $\hat{f}_{t-1}^{n}$ is defined as:
\begin{equation}
c^{n}(x_1, x_2)= \langle \hat{f}_{t}^{n}(x_1),\hat{f}_{t-1}^{n}(x_2) \rangle.
\label{eq:correlation}
\end{equation}

For computational efficiency, we only compute a partial cost volume with a limited search window.
Given a maximum displacement $D$ between two spatial locations $x_1$ and $x_2$, we compute a matching cost volume $C^n:\mathbb{R}^{w \times h \times (2D+1)^2}$, where $(2D+1)^{2}$ represents the number of correlations stacked along the channel dimension for every spatial location. Specifically,  $C^n$ is defined as:
\begin{equation}
  C^{n} = \{c^{n}(x, x + d)|x \in [1,w] \times [1,h], d \in [-D,D] \times [-D,D]\}.
\label{eq:displacement}
\end{equation}
Then, we apply \textit{argmax} to $C^n$ along the channel dimension to find the best matches to convert them into 2D displacement map $W^n$ as done in~\cite{Jaderberg2015}.
Finally, we warp $f_{t-1}^{N}$ and obtain a motion-compensated feature map $\tilde{f}_{t-1}^{N,n} = f_{t-1}^{N}(x+W^n)$.

The computational cost of computing $C^{n}$ with $D=10$ in our implementation is equivalent to that of a single convolution layer with $21\times 21$ kernels, while it does not include any learnable parameters.

\subsection{Effect of Maximum Displacement $D$}
The maximum displacement $D$ determines the search window size for our motion compensation network when feature matching is applied between temporal features (\cref{sec:mc_detail}).
Consequently, a larger $D$ leads the network to cover a large motion presented between frames, but it also increases the running time of the model.
\cref{tbl:effect_MC_D} compares the performances with different values of $D$, where both deblurring quality and computational time increases with $D$.
The results show that, for models with $D>10$, improvements in deblurring quality are marginal while increases in computational time are significant. 
We choose $D=10$ for our final models, which show much better deblurring quality compared to the model with $D=5$, while the running time of motion compensation is still fast.

\begin{table}[t]
\centering
\aboverulesep = 0.4mm %
\belowrulesep = 0.4mm %
\setlength\tabcolsep{0pt}
\caption{Effect of maximum displacement $D$. We compare models\protect\footnotemark[4] trained with different $D$, and measure deblurring accuracy on the DVD and GoPro datasets. MC time indicates the running time taken by motion compensation networks $\mathcal{M}^1$ and $\mathcal{M}^2$.}
\scalebox{0.80}{%
\begin{tabularx}{1.25\columnwidth}{>{\centering}p{0.3\columnwidth}YYYYYYY}
\toprule
\multirow{2}{*}{Datasets} & \multicolumn{4}{c}{PSNR/SSIM of our 2-stacked model with different $D$}\\
& $D=5$
& 10
& 20
& 30
\\
\midrule
DVD \cite{su2017deep} & 30.22/0.921 & 30.33/0.923 & 30.33/0.923 & 30.32/0.922 \\
GoPro \cite{nah2017deep} & 28.59/0.909 & 28.74/0.911 & 28.76/0.911 & 28.77/0.910 \\
\midrule[0.01mm]
Total time (ms) & 45 & 53 & 110 & 222 \\
MC time (ms) & 5 & 15 & 55 & 131 \\
\bottomrule

\end{tabularx}%
}
\label{tbl:effect_MC_D}
\end{table}
\begin{table}[t]
\centering
\aboverulesep = 0.4mm %
\belowrulesep = 0.4mm %
\setlength\tabcolsep{0pt}
\caption{Effect of feature matching skipping. We measure deblurring accuracy (PSNR/SSIM) on the DVD dataset~\cite{su2017deep} for $n$-stacked models\protect\footnotemark[1] with and without feature matching skipping. MC time indicates the running time of motion compensation networks.}
\scalebox{0.80}{%
\begin{tabularx}{1.25\columnwidth}{>{\centering}p{0.2\columnwidth}>{\centering}p{0.25\columnwidth}YYY}
\toprule
&& 2-stack
& 4-stack
& 10-stack
\\
\midrule
\multirow{3}{*}{\makecell{w/o\\skipping}}&\cellcolor{lightlightgray}PSNR/SSIM &\cellcolor{lightlightgray}30.33/0.923 &\cellcolor{lightlightgray}30.76/0.928 &\cellcolor{lightlightgray}31.09/0.933 \\
\cmidrule{2-5}
&Total time (ms) & 68 & 119 & 290 \\
&MC time (ms) & 30 & 60 & 152 \\
\midrule[0.02mm]

\multirow{3}{*}{\makecell{w/\\skipping}}&\cellcolor{lightlightgray}PSNR/SSIM  &\cellcolor{lightlightgray}30.33/0.923 &\cellcolor{lightlightgray}30.73/0.929 &\cellcolor{lightlightgray}31.07/0.933 \\
\cmidrule{2-5}
&Total time (ms) & 54 & 84 & 171 \\
&MC time (ms) & 15 & 16 & 19 \\
\bottomrule

\end{tabularx}%
}
\label{tbl:effect_corr_skip}
\end{table}
\footnotetext[4]{Note that the models are trained for 300K iterations.}

\subsection{Faster Motion Compensation with Feature Matching Skipping}
Although the computational overhead of our motion compensation network is small, it can be a burden if we stack a number of multi-task units.
We can effectively resolve this problem by skipping the feature matching operation in some stacks (\eg, $n>1$).
Specifically, we may use the motion pre-computed in an early stack for later stacks.
As the feature matching operation occupies most computations for motion estimation, 
we can save quite a large computations by skipping the operation.
Moreover, reusing the pre-computed motion still effectively boosts the deblurring quality, as our motion compensation network can produce moderate motion estimation results from early stacks (Figs 6e and 6f in the main paper).

\begin{table}[t]
\centering
\aboverulesep = 0.4mm %
\belowrulesep = 0.4mm %
\setlength\tabcolsep{0pt}
\caption{Quantitative evaluation on the REDS dataset~\cite{Nah2019REDS}. The colored dots denote models showing similar deblurring qualities in PSNR, where each of our models yields an upper bound PSNR at a much faster running time.}
\scalebox{0.9}{%
\begin{tabularx}{1.111\textwidth}{
Y
*{8}{>{\centering}p{0.11\textwidth}}
>{\centering\arraybackslash}p{0.11\textwidth}}

\toprule
& \multirow{2}{*}[-0.2em]{\makecell{EST$_{L}$\textcolor{cyan}{$^{\bullet}$}\\\cite{Zhong2020ECCV}}}
& \multirow{2}{*}[-0.2em]{\makecell{PVD\textcolor{cyan}{$^{\bullet}$}\\\cite{Son2021PVD}}}
& \multirow{2}{*}[-0.2em]{\makecell{PVD$_{L}$\textcolor{purple}{$^{\bullet}$}\\\cite{Son2021PVD}}}
& \multicolumn{4}{c}{Ours$^n$ ($n$: \texttt{\#} of stacks)} 
& Ours$^n_{hp}$ & Ours$^n_L$ \\
\cmidrule(lr){5-8} \cmidrule(l){9-10}
&  &  &  & $n=1$& 2 & 4\textcolor{cyan}{$^{\bullet}$} & 10\textcolor{purple}{$^{\bullet}$} & 2 & 10\textcolor{purple}{$^{\bullet}$} \\
\midrule
PSNR & 31.48 & 31.99 & 32.33 & 30.32 & 31.28 & 31.87 & 32.86 & 31.25 & \textbf{33.55} \\
SSIM & 0.922& 0.927 & 0.932 & 0.901 & 0.917 & 0.926 & 0.940 & 0.916 & \textbf{0.948} \\
\midrule[0.01mm]
Params (M) & 2.47 & 10.51 & 23.36 & \textbf{0.54} & 1.04 & 2.04 & 5.03 & 1.04 & 13.80 \\
Time (ms) & 132 & 238 & 585 & \textbf{33} & 53 & 86 & 171 & \textbf{33} & 416 \\
\bottomrule
\end{tabularx}%
}

\label{tbl:quanti_REDS}
\end{table}

\cref{tbl:effect_corr_skip} validates the effect of skipping feature matching on models stacked with different numbers of multi-task units.
Models without the skipping scheme perform correlation-based feature matching for every stack, while models with the skipping scheme reuse the motion computed by the feature matching operation in the first stack for the rest of the stacks.
As shown in the table, 
compared to the models that re-compute a motion in every stack,
the models skipping feature matching operations show slightly lower deblurring quality but record significantly reduced running time.

\begin{table}[t]
\centering
\aboverulesep = 0.4mm %
\belowrulesep = 0.4mm %
\setlength\tabcolsep{0pt}
\caption{Additional ablation study on our model with structure injection. Our structure injection consists of an addition operation of detail/structure features, and motion layer. The model in the last row is our final model, the same final model reported in the last row of Table 2 in the main paper.}
\scalebox{0.90}{%
    \begin{tabularx}{1.11111\columnwidth}{
    Y Y
    >{\centering}p{0.171\columnwidth}
    >{\centering}p{0.171\columnwidth}
    >{\centering}p{0.171\columnwidth}
    >{\centering\arraybackslash}p{0.171\columnwidth}
    }
    \toprule
    \multicolumn{2}{c}{\multirow{2}{*}{Structure injection}}& \multicolumn{2}{c}{\multirow{2}{*}{Deblurring quality}} & \multicolumn{2}{c}{\multirow{2}{*}{\makecell{Motion compensation \\accuracy}}} \\
    &&&&&\\
    \cmidrule{1-2}
    \cmidrule(l){3-6}
    addition & motion layer & PSNR & SSIM & PSNR & SSIM \\
    \midrule
     &     & 29.84 & 0.915 & 27.24 & 0.862 \\
    \rowcolor{lightlightgray} & \cm   &   29.84 & 0.916 & 27.24 & 0.862 \\
    \cm &     & 30.26 & 0.921 & 26.20 & 0.891 \\
    \midrule[0.01mm]
    \rowcolor{lightlightgray}\cm &\cm & 30.30 & 0.923 & 27.25 & 0.864 \\
    \bottomrule
    \end{tabularx}%
}
\label{tbl:structure_injection}
\end{table}
\subsection{Deeper Analysis on Structure Injection}
In our structure injection scheme, instead of using only the simple addition of detail and structural features, we further process the features with the motion layer for a more effective fusion of those features.
A question may naturally follow 
whether the motion layer is an essential component in the structure injection scheme.
To answer the question, we conduct an additional ablation study on our final model with components comprising the structure injection scheme, the addition operation between the detail-level and structural feature maps, and the motion layer that combines the feature maps (\cref{tbl:structure_injection}).
Note that, in the table, the model in the last row is the same final model used in the ablation study (Sec.~4.2.2) of the main paper.

In \cref{tbl:structure_injection}, using the motion layer without the addition operation of structural features (the second row of the table) does not have any effect other than slightly increasing the model size.
On the other hand, the addition of detail and structural features (the third row) brings a significant increase in the deblurring quality while retaining the motion compensation quality.
When the motion layer is attached (the last row), the deblurring quality further increases, thanks to the motion layer making the multi-task detail network more effectively focus on learning detail features by fully injecting the structural information into detail features.

\section{More Results}

\begin{figure}[t]
\centering
\def \wb {0.6} %
\begin{tabular}{@{}c@{}}
    \includegraphics[width=\wb\columnwidth]{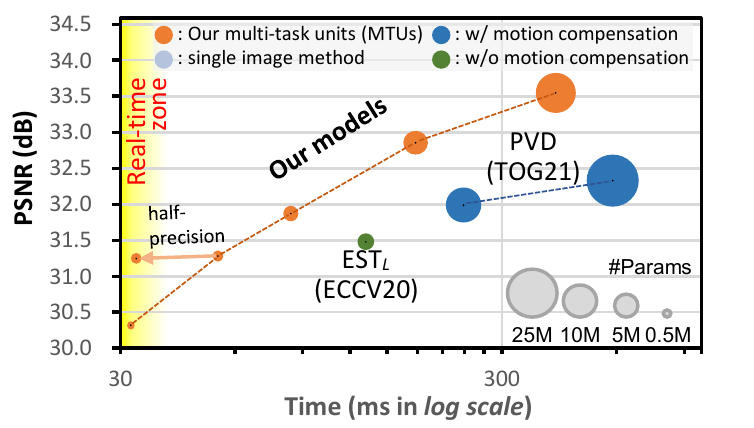}\\[-0.02in]
\end{tabular}
\caption{Comparison on deblurring efficiency. Our models are indicated in orange circles, each of which is stacked with a different number of multi-task units.}
\label{fig:efficiency_REDS}
\end{figure}
\input{data_supp/fig_object_detection}

\paragraph{Comparison on the REDS dataset}
In the main paper, we quantitatively compared our models with previous ones on the DVD dataset~\cite{su2017deep} and GoPro dataset~\cite{nah2017deep}. In this section, we provide an additional comparison on the REDS dataset~\cite{Nah2019REDS}.
Specifically, we compare our $n$-stacked models with previous video deblurring methods EST \cite{Zhong2020ECCV} and PVD~\cite{Son2021PVD} trained with the REDS dataset.
For the previous models, as the authors do not provide the model trained on the dataset,
we trained the models with the code provided by the authors to reproduce the same quantitative results reported by the authors.
For our models, we trained each model using the training set provided in the dataset, with the same training strategy described in Sec.~4.1 in the main paper.

\cref{tbl:quanti_REDS} shows quantitative results.
Note that EST$_L$ denotes the model with larger parameters compared to the EST model used in the comparisons on the DVD dataset~\cite{su2017deep} and GoPro dataset~\cite{nah2017deep} (Tables 3 and 4 in the main paper, respectively).
As the table indicates, compared to EST$_L$, our 4-stacked model shows better deblurring quality with a smaller model size and faster computation time.
Compared to PVD, our 4-stacked model reports comparable deblurring quality even with much smaller model size and faster running time.
Our 10-stacked model and the larger model (Ours$^{10}_L$) outperform PVD$_L$ by a large margin but still have a smaller model size and faster inference time, validating the effectiveness of our approach.

\cref{fig:efficiency_REDS} visualizes \cref{tbl:quanti_REDS}.
The diagram shows the similar tendency to the cases trained with the DVD~\cite{su2017deep} and GoPro~\cite{nah2017deep} datasets (Figs.~1a and 1b of the main paper),
where each of our model variants shows a much faster running time compared to previous methods with similar deblurring quality.

\paragraph{Application: Object Detection}
Thanks to the flexibility of our architecture, our network can cover from environments where high deblurring quality is desired to environments demanding for low-computation power.  For the latter case, our lightweight real-time model can be leveraged to pre-process videos to improve higher-level vision tasks such as object detection for autonomous driving, where real-time processing is highly demanded.

\cref{fig:object_detection} qualitatively shows the object detection results, for which we used our 2- and 10-stack models to deblur video frames of the GoPro dataset and applied object detection~\cite{Redmon2018YOLOv3AI} to the deblurred frames. Although the 10-stack model showed better deblurring results compared to the 2-stack model, the 2-stack model is enough to improve the object detection quality, despite its real-time speed.

\paragraph{Additional Qualitative Results}
In the main paper, we qualitatively compared our models with previous methods on the DVD dataset~\cite{su2017deep}, GoPro dataset~\cite{nah2017deep}, and real-world blurred videos~\cite{cho2012video}.
In this section, we provide qualitative results on the REDS dataset~\cite{Nah2019REDS} (\cref{fig:comparison_REDS_1,fig:comparison_REDS_2}).
We also additionally show qualitative results on the DVD dataset (\cref{fig:comparison_DVD_2}), GoPro dataset~\cite{nah2017deep} (\cref{fig:comparison_GoPro_2}), and real-world blurred videos (\cref{fig:comparison_real_2}).

\section{Detailed Network Architecture}
\label{sec:network_detail}
\Tbl{Network_Structure} shows our network architecture in detail.
Each multi-task unit (\textbf{MTU} in the table) consists of three main components:
\textit{multi-task detail network} $\mathcal{F}^n$,
\textit{deblurring network} $\mathcal{D}^n$,
and \textit{motion compensation network} $\mathcal{M}^n$,
where $n$ is the index of a multi-task unit.

The multi-task detail network $\mathcal{F}^{n}$ is a lightweight encoder-decoder network based on the U-Net architecture~\cite{Unet15}.
The encoder is composed of a convolution layer followed by two down-sampling convolution layers with stride two.
The decoder has two up-sampling deconvolution layers followed by a convolution layer.
The encoder and decoder are connected by skip-connections at the same levels.
Between the encoder and decoder, the network has four residual blocks.
The deblurring network $\mathcal{D}^{n}$ consists of a single convolution layer dubbed as a \textit{deblur layer} followed by an element-wise summation operator connected with a long skip connection carrying $I_t^b$ for the residual detail learning.
The motion compensation network $\mathcal{M}^n$ consists of a single convolution layer dubbed as a \textit{motion layer} followed by a motion compensation module having no learnable parameters.

\input{data_supp/fig_comparison_REDS_1}
\input{data_supp/fig_comparison_REDS_2}

\input{data_supp/fig_comparison_DVD_2}
\input{data_supp/fig_comparison_GoPro_2}

\input{data_supp/fig_comparison_real_2}
\begin{table}[t]
\centering

\aboverulesep = 0.0mm %
\belowrulesep = 0.0mm %
\caption{Detailed network architectures. \textbf{MTU}$^n$ stands for the $n$-th multi-task unit in the $N$-stacked \textbf{MTU}s, and \textbf{MC} means the motion compensation module. In the columns, \texttt{input}, \texttt{type}, \texttt{act}, \texttt{output}, \texttt{k}, \texttt{c}, \texttt{s}, \texttt{p} and \texttt{\#} denote the input, type, activation function, output, kernel size, out-channels, stride, padding, and repeating number of a layer, respectively. For the layer types, we have \textit{conv}, \textit{deconv}, \textit{identity}, and \textit{sum}, which denote convolution, deconvolution, identity, and element-wise summation layers, respectively. $[\cdot]$ indicates the concatenation operation in the channel direction.}
\scalebox{.8}{%
\begin{tabularx}{1.25\textwidth}{>{\centering}p{0.3\textwidth}>{\centering}p{0.13\textwidth}>{\centering}p{0.13\textwidth}>{\centering}p{0.13\textwidth}YYYYY}
\multicolumn{9}{c}{\textbf{Overall network architecture}}\\
\cmidrule[1pt]{1-9}
\rowcolor{lightlightgray}\texttt{input} & \texttt{type} & \texttt{act} & \texttt{output} & \texttt{k} & \texttt{c} & \texttt{s} & \texttt{p} & \texttt{\#} \\
\cmidrule[0.5pt]{1-9}
 
$I_t^b$ & \textit{conv} & \textit{relu} & $f^b_t$ & 3 & 26 & 1 & 1 & 1 \\
$[I_{t-2}^b \cdot I_{t-1}^b \cdot I_{t+1}^b \cdot I_{t+2}^b \cdot I_{t-1}^r]$ & \textit{conv} & \textit{relu} & $f_{t-1}$ & 3 & 26 & 1 & 1 & 1 \\
$[ f^b_t \cdot f_{t-1} ]$ \texttt{if} $n=1$ \texttt{else} $[f^{n-1}_t \cdot \tilde{f}^{N,n-1}_{t-1}]$ & $\textbf{MTU}^n$ & - & $[f^{n}_t \cdot \tilde{f}^{N,n}_{t-1}]$ & - & 52 & - & - & $N$-1 \\
$[ f^b_t \cdot f_{t-1} ]$ \texttt{if} $n=1$ \texttt{else} $[f^{n-1}_t \cdot \tilde{f}^{N,n-1}_{t-1}]$ & $\textbf{MTU}^N$ & - & $I^r_n$ & - & 3 & - & - & 1 \\
\cmidrule[1pt]{1-9}

\multicolumn{9}{c}{}\\
\multicolumn{9}{c}{}\\
\multicolumn{9}{c}{\textbf{Multi-tasking unit} $\textbf{MTU}^n$}\\
\cmidrule[1pt]{1-9}
\rowcolor{lightlightgray}\texttt{input} & \texttt{type} & \texttt{act} & \texttt{output} & \texttt{k} & \texttt{c} & \texttt{s} & \texttt{p} & \texttt{\#} \\
\cmidrule[0.5pt]{1-9}
$[ f^b_t \cdot f_{t-1} ]$ \texttt{if} $n=1$ \texttt{else} $[f^{n-1}_t \cdot \tilde{f}^{N,n-1}_{t-1}]$ & $\mathcal{F}^n$ & - & $f^n_t$ & - & - & - & - & 1 \\
$f^n_t$, $f^b_t$, $f^N_{t-1}$ & $\mathcal{M}^n$ & - & $[f^n_t \cdot \tilde{f}^{N,n}_{t-1}]$ & - & - & - & - & 1 \\
$[f^n_t \cdot \tilde{f}^{N,n}_{t-1}]$ & $\mathcal{D}^n$ & - & $I^{r,n}_t$ & - & - & - & - & 1 \\
\cmidrule[1pt]{1-9}
 
\multicolumn{9}{c}{}\\
\multicolumn{9}{c}{}\\
\multicolumn{9}{c}{\textbf{Multi-task detail network} $\mathcal{F}^n$}\\
\cmidrule[1pt]{1-9}

\rowcolor{lightlightgray}\texttt{input} & \texttt{type} & \texttt{act} & \texttt{output} & \texttt{k} & \texttt{c} & \texttt{s} & \texttt{p} & \texttt{\#} \\
\cmidrule[0.5pt]{1-9}
$[ f^b_t \cdot f_{t-1} ]$ \texttt{if} $n=1$ \texttt{else} $[f^{n-1}_t \cdot \tilde{f}^{N,n-1}_{t-1}]$ & \textit{conv} & \textit{relu} & conv$_1$ & 5 & 26 & 1 & 1 & 1 \\
conv$_1$ & \textit{conv} & \textit{relu} & conv$_2$ & 3 & 26 & 2 & 1 & 1 \\
conv$_2$ & \textit{conv} & \textit{relu} & res$_0$  & 3 & 26 & 2 & 1 & 1 \\
res$_0$  & \textit{identity} & - & skip & - & - & - & - & - \\
 
 \cmidrule(lr){1-9}
 res$_0$  & \textit{conv} & \textit{relu} & res$_{1-1}$  & 3 & 52 & 1 & 1 & \multirow{3}{*}{8} \\
 res$_{1-1}$  & \textit{conv} & \textit{relu} & res$_{1-2}$ & 3 & 52 & 1 & 1 &  \\
 res$_{1-2}$, res$_0$  & \textit{sum} & - & res$_0$ & - & - & - & - & \\
 \cmidrule(lr){1-9}
 res$_0$, skip & \textit{sum} & - & res & - & - & - & - & 1 \\
 res & \textit{deconv} & \textit{relu} & deconv$_1$ & 4 & 26 & 2 & 1 & 1 \\
 deconv$_1$, conv$_2$ & \textit{sum} & - & deconv$_1$ & - & - & - & - & 1\\
 deconv$_1$ & \textit{deconv} & \textit{relu} & deconv$_2$ & 4 & 26 & 2 & 1 & 1 \\
 deconv$_2$, conv$_1$ & \textit{sum} & - & $f^n_t$ & - & - & - & - & 1\\
\cmidrule[1pt]{1-9}

\multicolumn{9}{c}{}\\
\multicolumn{9}{c}{}\\
\multicolumn{9}{c}{\textbf{Motion compensation network} $\mathcal{M}^n$}\\
\cmidrule[1pt]{1-9}
 
\rowcolor{lightlightgray}\texttt{input} & \texttt{type} & \texttt{act} & \texttt{output} & \texttt{k} & \texttt{c} & \texttt{s} & \texttt{p} & \texttt{\#} \\
\cmidrule[0.5pt]{1-9}
$f^n_t$, $f^b_t$ & \textit{sum} & -  & sum$_0$ & - & - & - & - \\
sum$_0$ & \textit{conv} & \textit{relu} & $\hat{f}^n_{t}$& 5 & 52 & 4 & 1 & 1 \\
$\hat{f}^n_{t}$, $\hat{f}^n_{t-1}$, $f^N_{t-1}$& \textbf{MC} & -  & $[f^n_t \cdot \tilde{f}^{N,n}_{t-1}]$ & - & - & - & - \\
 \cmidrule[1pt]{1-9}
 
\multicolumn{9}{c}{}\\
\multicolumn{9}{c}{}\\
\multicolumn{9}{c}{\textbf{Deblurring network} $\mathcal{D}^n$}\\
\cmidrule[1pt]{1-9}
 
\rowcolor{lightlightgray}\texttt{input} & \texttt{type} & \texttt{act} & \texttt{output} & \texttt{k} & \texttt{c} & \texttt{s} & \texttt{p} & \texttt{\#} \\
\cmidrule[0.5pt]{1-9}
$[f^n_t \cdot \tilde{f}^{N,n}_t]$ & \textit{conv} & -  & $I^{res,n}_{t}$& 3 & 3 & 1 & 1 & 1 \\
$I^{res,n}_{t}$, $I^b_t$ & \textit{sum} & - & $I^{r,n}_{t}$ & - & - & - & - & 1 \\
 
\cmidrule[1pt]{1-9}
\end{tabularx}%
}
\label{tbl:Network_Structure}
\end{table}
\end{appendices}

\end{document}